# Hands-on experiments on intelligent behavior for mobile robots.

[a]Erik Cuevas, [a]Daniel Zaldivar[1], [a]Marco Pérez-Cisneros and [b]Marte Ramirez-Ortegon

[a]Departamento de Ciencias Computacionales
Universidad de Guadalajara, CUCEI
Av. Revolución 1500, Guadalajara, Jal, México
[b]Institut für Informatik, Freie Universität Berlin
Takustr. 9, 14195 Berlin, Germany

**Abstract**

In recent years, Artificial Intelligence (AI) techniques have emerged as useful tools for solving various engineering problems that were not possible or convenient to handle by traditional methods. AI has directly influenced many areas of computer science and becomes an important part of the engineering curriculum. However, determining the important topics for a single semester AI course is a nontrivial task, given the lack of a general methodology. AI concepts commonly overlap with many other disciplines involving a wide range of subjects, including applied approaches to more formal mathematical issues. This paper presents the use of a simple robotic platform to assist the learning of basic AI concepts. The study is guided through some simple experiments using autonomous mobile robots. The central algorithm is the Learning Automata (LA). Using LA, each robot action is applied to an environment to be evaluated by means of a fitness value. The response of the environment is used by the automata to select its next action. This procedure holds until the goal task is reached. The proposal addresses the AI study by (1) offering in LA a unifying context to draw together several of the topics of AI; and (2) motivating the students to learn by building some hands-on laboratory exercises. The presented material has been successfully tested as AI teaching aide in the University of Guadalajara's robotics group as it motivates students and increases enrolment and retention while educating better computer engineers.

## 1. Introduction

Artificial Intelligence (AI) techniques have emerged as useful tools for solving various engineering problems that were not possible or convenient to handle by traditional methods. The use of Artificial Intelligence within science is growing at a remarkable rate. In the early 1990s, little was known about how Artificial Intelligence could be applied in a practical way to the physical and engineering sciences. At that stage, few experimental scientists showed any interest in the area. Now, hundreds of research papers and applications are published each year and the numbers are rising rapidly. The change has been dramatic and yet, despite the growth, the field is still young. Hence, it is impossible to conceive an engineering formation without an AI course. On the other hand, it has been acknowledged that teaching an introductory course on Artificial Intelligence may be difficult for some students with little experience on heuristic topics [1, 2]. Several issues contribute to this predicament: breadth *vs* depth; formalist *vs* applied; the old and traditional *vs* the new and innovative way; the roles of philosophy, cognitive science, linguistics and psychology should play on, and so on [3].

Educational goals can be achieved over a wide range of costs. AI modelling and applications may be taught from paper-and-pencil exercises, traditional computer programming or hands-on robotics programming. Robotics itself allows the teaching of AI keeping a hands-on approach to motivate students because it represents a bridge among theoretical and practical aspects [7]. In addition, robotics allows students to debug their code easily since it can be tested in robotics exercises generating visual results. The benefits of hands-on robotics have been demonstrated repeatedly, such as in [4; 5; 6]. It is therefore not surprising that many universities are investigating the use of robotics plants to teach AI.

---

[1] Corresponding autor, Tel +52 33 1378 5900, ext. 7715, E-mail: daniel.zaldivar@cucei.udg.mx

Artificial intelligence encompasses methods for dealing with uncertain and unknown environments [8]. Although there are many approaches that search for the best solution to this kind of problems, most of them may not be applied straight to Robotics. On the contrary, the algorithm known as Learning Automata (LA) (see reference [9]), is impressively suitable to be applied to classical heuristic robot problems. In some sense, LA resembles an autonomous robot while learning, because it selects an action from a finite set and evaluates the outcome for a given unknown environment. Then, the response from the environment is used by the automaton to select the next action. By this process, the automaton learns asymptotically to select the optimal action. In particular, the way on which the automaton uses the response from the environment to select its next action is determined by the chosen learning algorithm.

On the other hand, the LEGO© robots offer interesting possibilities as robotic platform. They include relatively low-cost hardware and software suitable for implementing complex designs for Robotics, Mechatronics and Control. Students can become quickly familiar with the set while taking further steps by adventuring to the designing and testing their own add-on components. They might also interface the kit to other processing and sensing units such as microcontrollers, sensors, etc.

The main approach of this work is to assume that the solution of a classical AI problem may be conceived as robotics tasks, allowing its own implementation over a mobile robot platform. The proposal addresses the AI study by (1) offering in LA a unifying context to draw together several of the topics of AI; and (2) motivating the students to learn by building some hands-on laboratory exercises. The presented material has been successfully tested as AI teaching aide in the University of Guadalajara's robotics group as it motivates students and increases enrolment and retention while educating better computer scientists.

An autonomous robot is a system that is capable of acquiring sensory data and external stimulus in order to execute a given behaviours to affects its own environment. It decides for itself how to relate sensory data to its behaviour in order to attain certain goals. Such a system is able to deal with unpredictable problems, dynamically changing situations, poorly modelled environments or conflicting constraints. From these definitions, concepts from autonomous robots and learning automata seem to point to the same direction.

The paper presents four different experiments. Each test discusses on a basic AI concept. However, more complex principles –such as adaptation to dynamical environments, of AI may be easily designed on top of such simple experiments. All presented experiments are simulated in MatLAB&Simulink and implemented in real LEGO robots. For all them, the mapping between the environment signalling and its correspondent action to be taken is unknown and must be learned by the automata agent by means of a direct interaction with the environment. Such experience allows showing the effectiveness of learning in AI.

The paper is organized as follows: Section 2 develops the LA approach as it is used throughout the paper. Section 3 explains the mobile robot and its experimental setup while Section 4 discusses on each experiment and its results. Section 6 presents a wider view of the actual contribution of each experiment to the student understanding of AI. Some concluding remarks finish the paper.

## 2. Learning Automata

*2.1 Introduction*

LA operates by selecting actions via an stochastic process. Such actions operate within a given environment while being assessed according to a measure of the system performance. Figure 1 shows the typical learning system architecture. The automaton probabilistically selects an action (**X**). Such action is thus applied to the environment, while a cost function is calculated (*J*) and the performance evaluation function provides a reinforcement signal $\beta$. The automaton's internal probability distribution is updated whereby actions that achieve desirable performance are reinforced via an increased probability, while those actions yielding a decreased performance are penalized or left unchanged depending on the particular learning rule. It is assumed that over time, the average performance of the system will improve until one given limit is reached. Following common practices for optimization problems, the action with the highest probability would correspond to the global minimum as demonstrated by rigorous proofs of convergence available in [9] and [10].

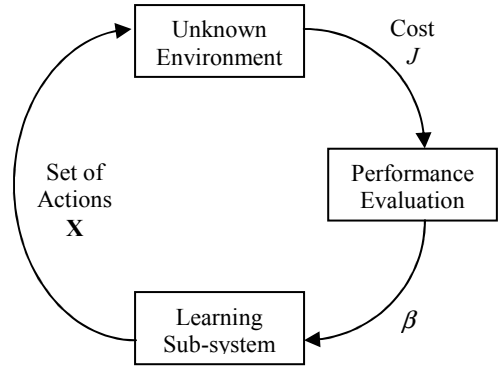

**Figure 1.** A simple reinforcement learning system

Learning Automata –also referred to as an 'automaton', is built over three key elements: a finite number of actions, an unknown environment, and a learning algorithm. An automaton can be described by a vector:

$$\mathbf{LA} = \{\boldsymbol{\alpha}, \boldsymbol{\beta}, F, \mathbf{p}\} \qquad (1)$$

where

$\boldsymbol{\alpha} = \{\alpha_1, \alpha_2, \ldots, \alpha_r\}$ is the set of action of the Automaton.

$\boldsymbol{\beta} = \{\beta_1, \beta_2, \ldots, \beta_r\}$ is the set of inputs to the Automaton, outputs of the environment.

$F$ is the function that maps current state and input into next state (learning algorithm).

$\mathbf{p} = \{p_1, p_2, \ldots, p_r\}$ is a set of the internal state, (probability distribution).

*2.2 Internal state and environment*

The internal state of the automaton $\mathbf{p}$ is represented by the action probabilities of the automaton. For mathematical simplicity, it is assumed that each internal state corresponds to a unique action. The input of the environment is one of the $r$ actions selected by the automaton. The output (response) of the environment to each action $i$ is given by $\beta_i$. When $\beta_i$ is a binary response (a Flag performance), the environment is said to be **P-Model** type. In such an environment, $\beta_i = 1$ is taken as failure while $\beta_i = 0$ is taken as success. In **Q-Model** environment, $\beta_i$ can take a finite number of values between [0,1], while in **S-Model** $\beta_i$ is a value between [0,1].

*2.3 Learning Algorithm in P-Models*

As shown in Eq. 1, the learning algorithm $F$ can be represented by:

$$\mathbf{p}(n+1) = F[\mathbf{p}(n), \boldsymbol{\alpha}(n), \boldsymbol{\beta}(n)] \qquad (2)$$

If operator $F$ is linear, the learning algorithm is said to be linear; otherwise, it is referred to as non-linear scheme. The fundamental idea behind all learning algorithms is as follows: if the LA selects an action $\alpha_i$ at iteration $n$ and obtains a convenient response from the environment, the action probability $p_i(n)$ is increased while the action probabilities of the other actions are decreased. For an unfavourable response, $p_i(n)$ is decreased, while the other action probabilities are increased. Thus, we have:

**Favourable response**

$$p_i(n+1) = p_i(n) + a[1 - p_i(n)]; \tag{3}$$

$$p_j(n+1) = (1-a)p_j(n); \quad \forall j; j \neq i \tag{4}$$

**Unfavourable response**

$$p_i(n+1) = (1-b)p_i(n); \tag{5}$$

$$p_j(n+1) = \frac{b}{r-1} + (1-b)p_j(n); \quad \forall j; j \neq i \tag{6}$$

with $a$ and $b$ representing the control parameters for the learning evolution. Their values falls within the interval [0,1]. The above equations give the general rule for the updating of the action probabilities. If in the above equations ($a = b$), a so-called $L_{RP}$ scheme is obtained, while ($b = 0$) results in the $L_{RI}$ scheme. Non-linear updating schemes have been pursued by researchers [11, 12], but no significant improvement over the linear updating schemes has been reached. A crucial factor that limits applications involving LA is their slow rate of convergence. This factor becomes more pronounced when the number of actions increases and the LA has to update a greater number of action probabilities.

*2.4 Learning Algorithm in S-Models*

Since the response from the environment for the case of the S-Model is a random variable between [0.1], its application to learning system problems requires *a priori* knowledge of the lower and upper bounds of the performance indexes in order to scale the responses between [0,1]. The updating rule for the *S* scheme is as follows: suppose that for a given iteration n, action $\alpha_i$ is chosen and the response from the environment is $\beta$, yielding:

$$p_i(n+1) = p_i(n) + a(1-\beta)(1-p_i(n)); \tag{7}$$

$$p_j(n+1) = p_j(n) - a(1-\beta)p_j(n); \quad \forall j; j \neq i \tag{8}$$

considering that the learning parameter falls within $0 < a < 1$.

## 3. Experiments

*3.1 Preliminaries*

The experiments aim to enable the student to solve one of the most important paradigms in AI, learning. The overall experience is conducted over the LEGO robotic platform. For the behaviour, a piece of code to learn control strategies is thus implemented following the LA principles previously discussed.

The mobile unit used throughout this paper is presented by Figure 2(a). The robot is based on the LEGO© NXT system. The autonomous robot refers to the differential mobile robot architecture which is characterized on Figure 2(b) following a state-space notation as follows:

$$\begin{bmatrix} \dot{x} \\ \dot{y} \\ \dot{\theta} \end{bmatrix} = \begin{bmatrix} -(c\sin\theta)/2 \\ (c\cos\theta)/2 \\ -c/b \end{bmatrix} \omega_l + \begin{bmatrix} -(c\sin\theta)/2 \\ (c\cos\theta)/2 \\ c/b \end{bmatrix} \omega_r \tag{9}$$

Here the coordinates of the back shaft's centre are defined by $(x, y)$, with $\omega_l$ y $\omega_r$ being the left and right wheel velocities respectively, $\theta$ representing the robot's orientation, and $b$ being the length of the back shaft. Following this configuration, $c$ represents the wheel's radio which is the key parameter in the model –a wider reference for this model may be found in [13]. For de simulation of the Eq. (9) in the experiments, MatLab&Simulink© are used as the developer platform, thanks to its well-known conveniences –such as a comprehensive literature, for teaching and training in Engineering.

On the other hand, although different languages and compilers for the LEGO© platform are currently available, in the real experiments, the ROBOTC© compiler is chosen as a convenient platform due to its trade of between real-time performance and programming easiness.

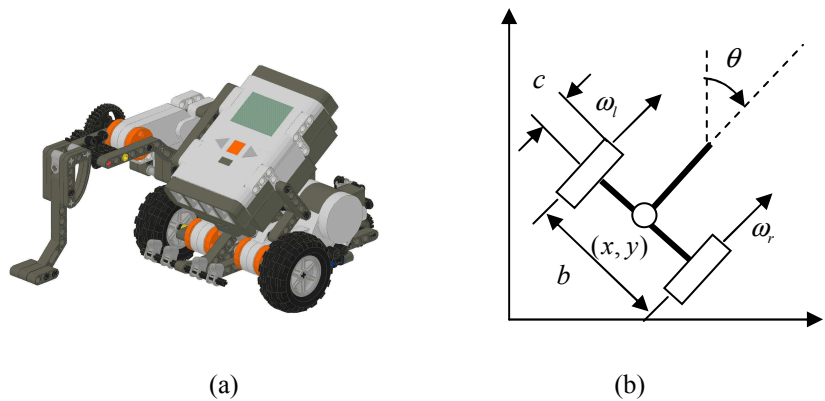

(a)          (b)

**Figure 2.** (a) LEGO platform and (b) Differential kinematics model.

*3.2 Experimental setup.*

The first step in the experiments is to create a set of actions (movements) that can be executed by the robot architecture. The action is considered using the mechanical limitations (non-holonomic constraints) of the robot presented in Fig. 2(a) and 2(b). The movements are forward, left forward, right forward, backward, left backward and right backward (see Fig. 3).

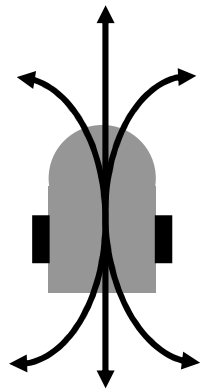

**Fig.3.** Directions of movement for the mobile robot

The differential model is manipulated changing the left and right wheel velocities. Making a combination on which wheel should be activated and the way the wheels turn, the action necessary for the test can be easily created as shown by Table 1. The velocities on the wheels are represented by $\omega_r$ for the right wheel velocity and $\omega_l$ for the left wheel velocity.

| Action | Direction | Right wheel | Left wheel |
|--------|-----------|-------------|------------|
| 1 | Forward | $\omega_r$ | $\omega_l$ |
| 2 | Right forward | 0 | $\omega_l$ |

| 3 | Left forward | $\omega_r$ | 0 |
| 4 | Backward | $-\omega_r$ | $-\omega_l$ |
| 5 | Right backward | 0 | $-\omega_l$ |
| 6 | Left backward | $-\omega_r$ | 0 |

**Table 1.** Mapping between actions and inputs of the differential model

For convenience $\omega_r$ and $\omega_l$ are considered to have the same value, so turning at every direction will generate the same deviation degree. To start the each experiment, the same probability in every action (1/6 each) is considered. The initial position of the robot is field's centre that is: $x$=0 and $y$=0. The goal point is calculated randomly to ensure that the algorithm is able to calculate a good response at any circumstance.

*3.3 The LA P-model for the robot.*

The whole algorithm may be divided into four sections: initialization, impact measurement, adaptation of the probability distribution and action selection. Figure 4 shows the full algorithm as coded for MatLab©, which it is assumed to be easy to be followed. However, a translation from Fig. 2 to ROBOTC code results straightforward. The robot operation, according to the algorithm, may be summarized as follows:

3.3.1 Initialization

Six feasible actions share the same probability of being chosen, i.e. a uniformly distributed probability (see initialization from Fig. 4).

3.3.2 Impact measurement

The chosen action, which was lately executed by the automaton, modifies the current robot's position. This evidently may have reduced or increased the distance between the mobile robot and the target point. To keep experiments as simple as possible, they only consider a flag-value to evaluate the quality of such action. Such flag is computed as follows:

$$\beta_{FL} = \begin{cases} 0 & \text{if } d(n) \geq d(n-1) \\ 1 & \text{otherwise} \end{cases} \quad (10)$$

with $d(n)$ y $d(n-1)$ being the actual distance and the last robot's position with respect to the target point.

3.3.3 Adaptation of the probability distribution.

By using Equations (3)-(6), the last distribution and the last flag value, it is possible to calculate the new value for the probability distribution (see measurement of the impact on the environment from Fig. 4).

| Initialization | `Pr=ones(1,6)*1/6;` |
|---|---|
| Measurement of the impact on the environment. | $\beta_{FL} = \begin{cases} 0 & \text{if } d(n) \geq d(n-1) \\ 1 & \text{otherwise} \end{cases}$ |
| | ```
if (β_FL=0)
    for val=1:6
    if(val==ac)
        Pr(ac)=Pr(ac)+a*(1-Pr(ac));
``` |



| | |
|---|---|
| Adaptation of the probability distribution | ```
            else
                Pr(val)=(1-a)*Pr(val);
            end

        end
    else
        for val=1:6
        if(val==ac)
            Pr(ac)=(1-b)*Pr(ac);
        else
            Pr(val)=(b/5)+(1-b)*Pr(val);
        end

        end
    end
``` |
| Action selection | ```
z=rand;
    sum=0;
for ind=1:length(Pr)
sum=Pr(ind)+sum;
  if (sum>=z)
     a=ind;
     break
end
end
``` |

**Fig. 4.** MatLab© code for the LA algorithm in all the experiments.

The action selection step considers a discrete distribution. A random uniformly distributed number $z(n)$ is first chosen. The probabilistic values for each of the actions are then added into an accumulative sum, until its value is greater or equal to the random value $z(n)$, yielding:

$$\sum_{i=1}^{ac} p_i(n) \geq z(n), \quad (11)$$

with *ac* representing the index of the selected action (see the implementation in Action selection from Fig. 4).

*3.4 Simulation*

In order to implement all experiments, the algorithm 3.3 must be adapted to the model described by Equation 9. Both are assembled into a Simulink© model which from now on will be identified such as platform, as indicated by Figure 5. The operation values on the simulation were chosen according to a differential LEGO mobile robot.



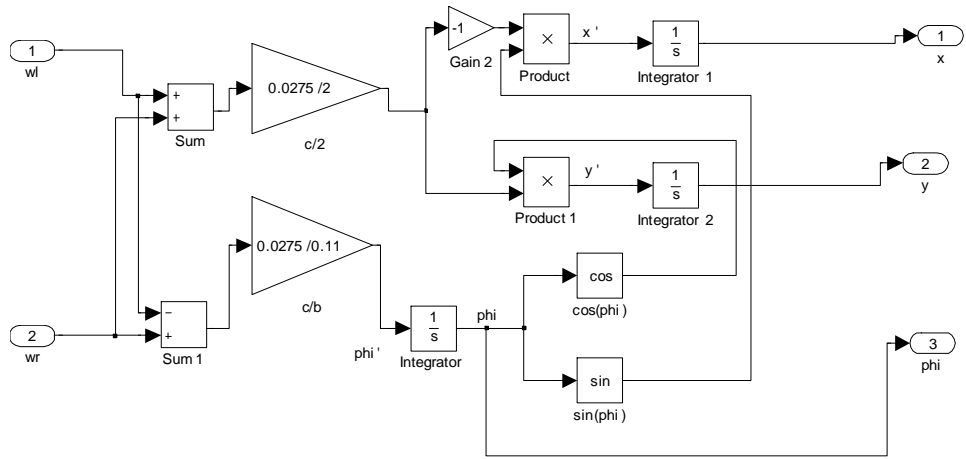

**Fig. 5.** Platform Simulink© model.

*3.5 Testing the experiment.*

Four experiments are presented in this paper. First, the parameters of the Eq. (3)-(6) are set to *a*=0.7 and *b*=0.7. Under this configuration, the learning system just awards the system each time a 'good' action is completed, i.e. the robot's position error is reduced. A bad action is also punished with same severity.

The second experiment applies an increased awarding scheme, considering no punishment for failures, i.e. *b* is zero and *a*=0.7.

On the other hand, the third experiment put a strong attention into punishment by taking the value of *b*=0.7 and making *a* zero.

The last experiment tests *a*=0.7 y *b*=0.7. However, obstacles are laid in the robot's workspace to force a more demanding learning problem because the robot will not be able to transit over those places.

The experimental set gathers basic tests that offer valuable insights into the learning process. However, it is pretty easy to add more demanding and complex tests. For instance, experiments regarding the automatic adaptation of values *a* and *b* in order to improved the overall robot response.

For all experiments, it is considered that the robot has reached the objective if it is a ±2 centimetres interval.

## 4. Results

This section discusses results from the four experiments; all the scales refer to centimetres. Figure 6 presents the robot's behaviour after testing the first experiment. The Figure show the simulation and the performance of the real robot. It is quite evident how the automaton in both cases reaches rather quickly the goal position as a result of receiving a correction value for both states, successful and failure. Figure 7 shows the result after Experiment 2 is performed. In this case, rewarding is stronger with values *b*=0 and *a* =0.7 and it is more difficult to reach the objective as a result. Figure 8 shows the result of experiment 3. Just as it is shown by the experiment 2, when the automaton corrects only for one action, it is more difficult to reach the objective position. Figure 9 shows the automata performance when awards and punishment values are used, but considering two new obstacles on the robot's path. The experiment shows how the automaton is able to avoid both obstacles but it is also evident that the learning environment as become more restrictive.



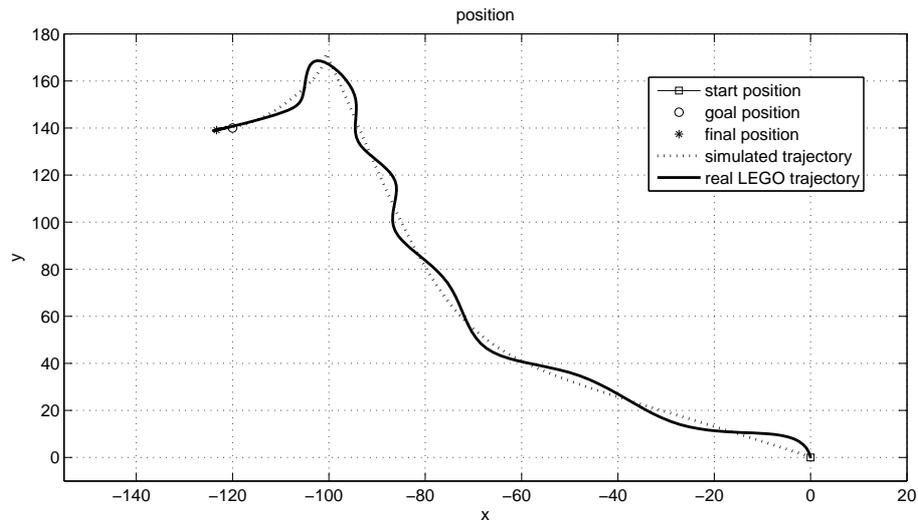

**Fig. 6.** Results from Experiment 1 considering *a*=0.7 and *b*=0.7.

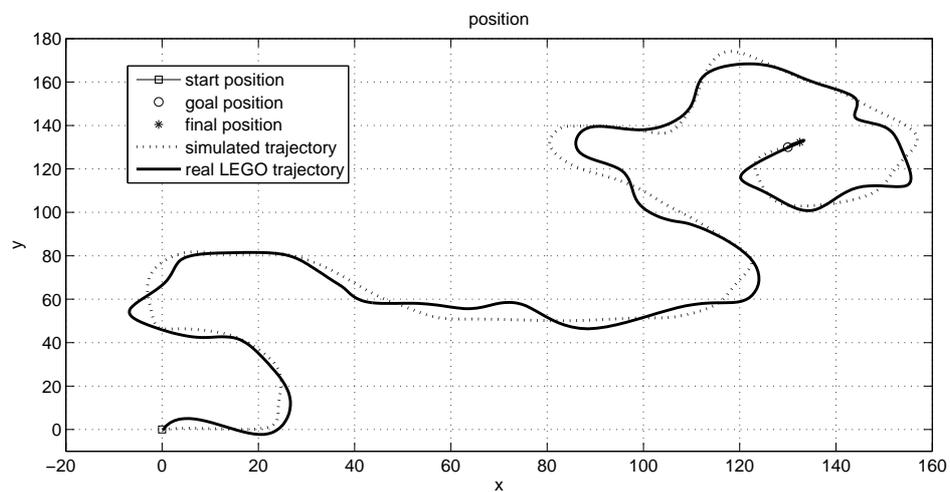

**Fig. 7.** Results from Experiment 2 with values *a*=0.7 and *b*=0.

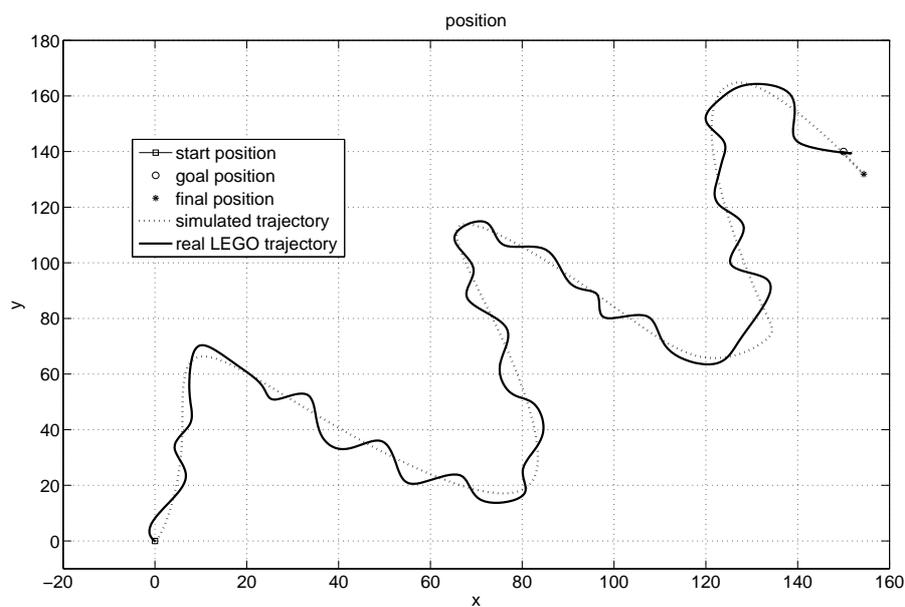

**Fig. 8.** Mobile robot trajectory as a resulting from Experiment 3, with *a*=0 y *b*=0.7, i.e. only punishment is applied.



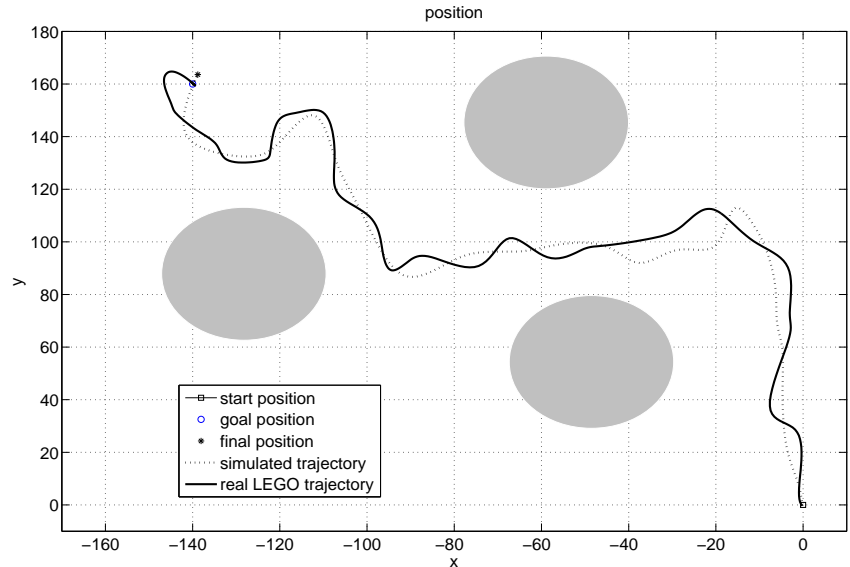

**Fig. 9.** Two obstacles are located in the robot's way while punish and rewards signals are applied in Experiment 4 with values *a*=0.7 and *b*=0.7.

En todos los casos presentados (Fig. 6-9) resulta evidente una clara diferencia presentada entre el comportamiento del autómata en simulación y el experimentado en el robot LEGO real. Dichas inconsistencias se deben al ruido provocado en la evaluación de la Ecuación 9 (Measurement of the impact on the environment) producto de la incertidumbre producida en la lectura de los encoders utilizados para medir la posición del robot en el plano. En tanto que en el caso de simulación, la distancia del robot presenta el caso ideal determinado por un integrador libre de ruido, producido por patinadas o frenadas realizadas por el robot real.

## 5. Conclusions

Considering that Artificial Intelligence has directly influenced many areas of computer science and becomes an important part of the engineering curriculum, this paper presents a solution to the difficulties of structuring pertinent knowledge in the teaching of the Artificial Intelligence.

The paper through a set of experiments introduces main concepts of Artificial Intelligence. The experiments are simple and easy to follow aiming to provide an easy to follow understanding of basic reinforcement learning concepts such as reward and punishment.

The platform (LEGO robot and the LA algorithm) profits from similarities between LA and an autonomous robot to support teaching about AI concepts. The results may be easily explored as the robot's behaviour is visually evident avoiding abstract and long mathematical developments to introduce AI concepts.

The four experiments show basic behavioural levels for AI. However, they may be easily modified to implement more advanced AI features.

All the experiments have been successfully tested over several courses and seminars on the topic in the University of Guadalajara, Mexico. Students coming from several backgrounds have recognized an easy approach of the experimental set to support the learning of basic AI concepts.